\newif\ifisfinal\isfinaltrue
\ifisfinal%
\documentclass[conference]{IEEEtran}
\else
\documentclass[peerreview]{IEEEtran}
\fi

\usepackage[init=false]{calsum}
\usepackage{zlhref}
\usepackage{zlsec}
\usepackage[ref]{zlfig}
\usepackage{zlreport}
\usepackage{zledu}
\usepackage{zlmath}
\usepackage{dblfloatfix}
\usepackage{eso-pic}
\AddToShipoutPictureFG*{%
  \AtPageLowerLeft{%
    \raisebox{12mm}{%
      \hspace*{\dimexpr 1in+\hoffset+\oddsidemargin\relax}%
      \parbox{\textwidth}{%
        \scriptsize\raggedright
        © 2026 IEEE. Personal use of this material is permitted. Permission from IEEE must be obtained for all other uses, in any current or future media, including reprinting/republishing this material for advertising or promotional purposes, creating new collective works, for resale or redistribution to servers or lists, or reuse of any copyrighted component of this work in other works.\\[1mm]
        DOI: 10.1109/ICHI69079.2026.00023
      }%
    }%
  }%
}

\calsumdocvar%
\usepackage[bibname=calsum]{zvieee}
\zaveninit%
\newcommand{\eauth}[3]{%
  \IEEEauthorblockN{#1}
  \IEEEauthorblockA{\textit{#2} \\
    \textit{\theorganization} \\
    Chicago, USA \\
    #3}}
\renewcommand{\zavenauthors}{
\author{%
\eauth{\theauthor}{\theorganizationdept}{\theauthoremail}
\and
\eauth{Sitara Rao}{College of Medicine}{srao11@bidmc.harvard.edu}
\\
\eauth{Barbara Di Eugenio}{\theorganizationdept}{bdieugen@uic.edu}
\and
\eauth{Aaron Chaise}{College of Medicine}{aaron.chaise@aah.org}}}
\hypersetup{pdfauthor={\theauthor; Sitara Rao; Aaron Chaise; Barbara Di Eugenio}}

\tabvarsize[\tabsize]{.85}
\renewcommand{\figrefname}{\mbox{Figure}}

\begin{document}
\zavenbegindoc%

\begin{abstract}
  Discharge summaries are lengthy medical documents that summarize a hospital
  in-patient visit.  Automatically generating them can reduce documentation
  burden and return clinician time to patient care. Whereas \Acfp{llm} could be
  used for this task, their Achilles heel is hallucinations, which can have
  drastic consequences for clinical documentation.  We present an
  evidence-driven alignment framework for discharge summarization at the
  clinical encounter level, that treats provenance as a first-class constraint,
  using semantic graphs and deep learning models. Each summary sentence is
  selected and organized via cross-document semantic alignment and is
  accompanied by explicit evidence links to its source spans.  We show our
  results on two corpora: a publicly available corpus (\acs{mimic}) and
  clinical notes written by physicians at the \uihealth.  Additionally, we make
  source code and trained models available.

\end{abstract}
\begin{IEEEkeywords}
\textit{summarization, EHR, graph, AMR\@.}
\end{IEEEkeywords}
\acresetall% reset the "memory" of acro

\sect{Introduction}

Automatic summarization is the task of using computers to summarize natural
language
text~\cite{maynezFaithfulnessFactualityAbstractive2020,ranjithaAbstractiveMultidocumentSummarization2017,erkanLexRankGraphbasedLexical2004,zhuLeveragingSummaryGuidance2023}.
This task has developed across many areas in computational linguistics for more
than six decades~\cite{luhnAutomaticCreationLiterature1958}.  Recently \llms\
have been
influential~\cite{brownLanguageModelsAre2020a,wangSelfConsistencyImprovesChain2022,huLoRALowRankAdaptation2021a}
and shown to achieve \sota\ performance in
summarization~\cite{lewisBARTDenoisingSequencetoSequence2020,zhangPEGASUSPretrainingExtracted2020}.
However, given their memory constraints, these models cannot condense a large
number of long documents
~\cite{xuWhenDoesDivide2025,liuLostMiddleHow2024,anilGeminiFamilyHighly2024}
and lack explainability~\cite{zhangOptimizingFactualCorrectness2020,\dsprovct}.
Hallucinations (erroneous and nonfactual generated text) present additional
challenges with \llm\
summarizations~\cite{kalaiWhyLanguageModels2025,iaroshReducingFactualHallucinations2025,yehudaInterrogateLLMZeroResourceHallucination2024}.

Our goal is to automatically generate an English \ds\ using clinical notes upon
discharge of a patient from a hospital.  A \ds\ is a medical document that
explains a patient's illness, reason for their hospital stay, and treatment.
Clinicians write exhaustive time-consuming documentation during hospitalization
of their patients, which can be aided by an automatic process.
\figref{note-flow} illustrates the correspondence (provenance) between
previously written clinical notes (note antecedents), stored in the \ehr\
system, and corresponding components of the \ds.

\introNoteFlowFig[t!]% -- intro note flow image

Clinical summarizations must be both faithful and
traceable~\cite{shingClinicalEncounterSummarization2021,zhangOptimizingFactualCorrectness2020}.
Generating \dss\ with these requirements highlights the difficulty of the task
for admissions of extended stay patients, which have \ac{ehr} notes that can
number in the thousands\footnote{Each admission's notes include those written
  during a single hospital stay.}.  Contemporary \sota\ methods, such as
fine-tuning \acp{llm}, render the task nearly impossible with the volume of
information for patients with prolonged hospitalizations.  In some cases,
\acp{llm} might feasibly summarize clinical documentation on a per note basis.
However, total textual content across all free text notes of admissions easily
extends past the limit of \acp{llm}' context window.  Even the impressively
large 2-million-token window of Gemini 2.0~\cite{anilGeminiFamilyHighly2024} is
not large enough for lengthy admissions\footnote{The largest admission includes
  1,233 notes in the \mimicname\ corpus.}.

Because of these memory
constraints~\cite{yehudaInterrogateLLMZeroResourceHallucination2024}, other
solutions are needed for the large multi-document automatic summarization task
of generating the \ds.  Abstractive methods can exhibit reduced faithfulness,
since they attempt at generating new text, sometimes by paraphrasing, rather
than directly copying source text, as in extractive
summarization~\cite{kryscinskiEvaluatingFactualConsistency2020,maynezFaithfulnessFactualityAbstractive2020}.
They also provide no traceable means of cross-referencing the text of
summarized documents.  We chose an extractive formulation with explicit
evidence linking with note section coupling so outputs remain auditable under
encounter-level multi-document inputs.

Previous methods have shown success at summarizing a single
section~\cite{adamsWhatSummaryLaying2021}, or one category such as physician
notes~\cite{gaoSummarizingPatientsProblems2022}.  However, to the best of our
knowledge, no prior work has combined episode-level clinical encounter
multi-section discharge summarization with explicit sentence-level provenance
evidence mapping.  This motivates the extractive methods formulated in this
work and provides a baseline for future abstractive summarization.

The contributions of this work include
\begin{zlenumerateinline}
\item an encounter--level,\footnote{An encounter is a hospitalization
    or visit summarized as a single event.} multi-document discharge summarization pipeline
\item a provenance-centered evaluation strategy aligned with sentence-level
  evidence links from summary statements to source spans
\item an evidence-linking constraint treated as a first-class output
\item a section-aware organization method induced by the Source Section Model
  \seesec{smy:srcsecds}
\item reusable \calsumurl[source code] to reproduce our results
\item \mimic\ generated discharge summaries with physician qualitative
  evaluations
\end{zlenumerateinline}
 % includes abstract
\sect[related]{Related Work}

Summarization is a well-established area in
\ac{nlp}~\cite{\liuct,\liaoct,erkanLexRankGraphbasedLexical2004,luhnAutomaticCreationLiterature1958}
that also includes work in the clinical
domain~\cite{zhangOptimizingFactualCorrectness2020,raffelExploringLimitsTransfer2020}.
The literature is rich with examples of clinical note summarization that
include both longitudinal~\cite{hirschHARVESTLongitudinalPatient2015}, and
non-longitudinal~\cite{pivovarovAutomatedMethodsSummarization2015} note types,
two examples of mutual discipline interest.  Furthermore, the shared
understanding, agreement, and acknowledgment that faithful summarization is
necessary, but lacking, has been thoroughly
reviewed~\cite{zhangOptimizingFactualCorrectness2020} despite recent efforts to
ameliorate hallucination using graph
methods~\cite{iaroshReducingFactualHallucinations2025}.  Canonical extractive
summarization methods have focused on ranking or selecting sentences from the
source text. Classic unsupervised approaches used graph-centrality ranking to
identify salient sentences~\cite{erkanLexRankGraphbasedLexical2004}. More
recent extractive methods leverage neural architectures to model cross-sentence
relationships in a unified graph and classify sentences for extraction, as
proposed by \citet{wangHeterogeneousGraphNeural2020}.

\Ac{amr} is \amrdef~\cite{\banarescuct}.  Recent approaches that use \ac{amr}
models as the primary data representation include work in
\ac{nlg}~\cite{manningHumanEvaluationAMRtoEnglish2020},
\acl{mt}~\cite{blloshmiXLAMREnablingCrossLingual2020}, and
\acl{qa}~\cite{limKnowWhatYou2020}.  The well-known work of \citet{\liuct} used
reduction methods with \ac{amr} graphs for summarization.  In this work, the
authors created a fully connected graph that was used heuristically to generate
abstractive text.  This was later broadened with a more comprehensive and
robust \ac{amr} graph based realization algorithm for multi-document
summarization~\cite{ogormanAMRSentenceMultisentence2018,\liaoct}.

Our work was inspired by the work of \citet{\liuct} and \citet{\liaoct} on
\ac{amr}-graph reduction.  Our method differs in that it learns and supervises
cross-document alignments using a network-flow
method~\cite{gaoFullyDynamicElectrical2022}.  These sentence-level provenance
links from summary statements to source spans provide the traceability clinical
summarization demands.  In contrast, prior graph-reduction work built on
sentence compression techniques~\cite{thadaniSentenceCompressionJoint2013},
such as re-framed commodity-flow~\cite{magnantiOptimalTrees1995}, which selects
edges for inclusion within a single sentence rather than cross-document
linking.  To the best of our knowledge, no prior work has combined \amr\ graphs
with flow-based (network-flow) alignment for \ds\ generation; while at the same
time, ensuring traceability by inferring explicit sentence-level provenance
links to source spans.

Several recent works report abstractive and LLM-based results on MIMIC-derived
discharge-summary generation
tasks~\cite{zhuLeveragingSummaryGuidance2023,palNeuralSummarizationElectronic2023}.
However, our work is not directly comparable because they differ in input scope
(single-note or fixed subsets vs.\ encounter-level multi-document aggregation),
target definition (full \ds\ vs.\ specific sections), preprocessing, and
evaluation pipelines, and they do not enforce explicit sentence-level
provenance as an output requirement.

\sect[smy:method]{Methods}

Our approach consists of a pipeline of six steps, as illustrated in
\figref{pipeline-overview}. Two corpora, the \mimiccite\ corpus and the
\acf{uicds}, which we collected in-house, were used for all experiments.  The
\uicds\ is an IRB-approved private dataset of 11,001 admissions and 607,872
notes, which include progress notes, radiology, ECG and additional note
categories from the \uihealth\ hospital. We will now discuss the modules in
\figref{pipeline-overview} in detail.

\pipelineOverviewFig[t] % pipeline-overview -- pipeline overview

\subsect[medsecid]{Section Notes by Clinical Topic}

The \spacyurl\ and \scispacyurl\ libraries were used to tokenize, sentence
chunk and tag biomedical and non-biomedical named entities from clinical text
input.  MedCAT~\cite{kraljevicMultidomainClinicalNatural2021} was used to link
tokens to \umls\ \acfp{cui} that aid in graph aligning their text-to-graph
concepts.  The parsed text was then segmented by clinical topic using the
\medsecid\ baseline model without modification.  \medsecid\ is a dataset of
clinical section annotations from the \mimic\ corpus across five note types and
50 sections on which we developed this model.

\subsect[smy:method:admgraph]{Construct Admission Graph}

A patient is admitted to the hospital upon entering for any administered
healthcare services.  From the healthcare perspective, this admission includes
what is done to the patient for the duration of the hospital stay:  the
admission graph is a semantic representation of it.  It is composed of two
disconnected graph components: all the antecedent notes for the admission and
the \ds.

After notes are sectioned, they are then parsed into graphs using the clinical
\amr\ \ac{thyme}.  Antecedent \amr\ graphs for the various notes are connected
to create the source graph and the \amr\ graphs of the \ds\ form the summary
graph.  These two disconnected components follow the structure of the source
and summary components of the bipartite graph we described in
\calamr~\cite{\plcalamrct} and in further detail in \secref{meth:align}.
However, document nodes that represent note categories, note sections and
clinical text paragraphs are used between the roots (as a ``has a'' relation)
and their respective \amr\ subgraphs as shown in \figref{adm-graph}.  Afterward,
we joined the \ac{amr} parser's selected sentence roots to their corresponding
paragraph nodes and used Coreference Resolution in place of concept node
merging to avoid loss of data.

\admGraphFig[t] % adm-graph -- admission graph figure

\subsect[meth:align]{Align Source and Summary}

\calamr\ (\calamrlongname) was leveraged to find clinical notes and candidate
sentences to use for summarization.  We refer the reader to our previous
work~\cite{\calamrct}, but we give a brief overview here.  These two graphs
start as separate components that become one bipartite graph.

Nodes are connected, as bipartite edges, if their semantic similarity's
neighborhood exceeds a threshold.  This similarity measure is calculated based
on embeddings assigned to concept and attribute \ac{amr} nodes and
\propbank~\cite{\propbankct} roles and role set edges.  The similarity measures
are also used as the information gain across the connected graph and all
subgraphs of each of the two in the max flow
algorithm~\cite{gaoFullyDynamicElectrical2022,ford1962flows}.  The assigned
flow values to each bipartite edge lead to the ``starvation'' of low
information subgraphs.  Subgraphs are effectively removed by setting low flow
alignment edge capacities to zero.

The novelty of our approach lays in clinical encounter--level supervision and
organization guided by \calamr\ alignments.  We used a network-flow
formulation~\cite{ford1962flows} to induce sentence-level alignment links
across encounter documents; these links serve both as supervised training
pairs and as explicit evidence-to-source span provenance for traceability.

\graphStatMergeTab[b]{\tabsize}

A total of 3,520 of the 11,957 \mimic\ admissions were aligned with \ac{calamr}
to create the Source Section Dataset~\seesec{smy:srcsecds}.
\Tabref{graphStatMergeTab} shows the average number of alignments across note
antecedent and \ds\ components and the average number of \acp{reentrancy}
(\reentrancydef) per admission~\cite{\calamrct}.  The ``alignable'' statistics
are nodes that are alignment candidates, such as concept and attribute notes.
The ``aligned'' statistics are those nodes with alignment edges.

Processing very large admissions (up to 1,233 notes in \mimic) resulted in
performance issues given the large scale of text.  Of the \uicds\ sample of
607K notes, the evaluation used an aligned subset of 834 notes due to performance
issues given the large scale of text \seesec{sec:lim}.

\subsect[smy:sentmatch]{Match Sentences}

Aligning the \uicds\ resulted in additional challenges.  The dataset has more
notes across category types compared to \mimic\ because the latter only
includes \ac{icu} notes~\cite{\pldsprovct}.  The consequence of this more
robust note variety is that admission note counts are much higher, and
therefore, take much longer to align.  There is also a higher risk of missed
alignments due to a potentially higher rate of reentrancies, which lead to flow
issues~\cite{\plcalamrct}.  Even though the \mimic\ alignments far outnumber
the \uicds, the \uicds\ has many more reentrancies.

The Sentence Matching Algorithm uses the \calamr\ alignments to identify the
sentences that best represent the summary.  This classification is based on the
sentence-to-sentence information gain from the aligned graph flow network.
\figref{adm-match} shows how the source sentence, \textit{Pre-cardiac
  catheterization assessment}, matches with the \ds\ sentence
\textit{``Coronary artery disease, status post coronary artery bypass
  grafting,''} by creating paths through the graph from a source sentence to a
summary sentence.  Each sentence connected in this way becomes a candidate.

\admMatchFig[t] % adm-match -- sentence matching algorithm

\noindent
The Sentence Matching Algorithm follows:
\begin{zlpackedenum}
\item For each \ds\ sentence in the reduced graph~\cite{\calamrct}, use a
  depth-first search to index aligned nodes
  (\figrefsub{adm-match}{a}).\label{step:smy:sentmatch:dfs}
\item For each indexed node in step~\ref{step:smy:sentmatch:dfs}, traverse the
  alignment edge to source nodes in the note antecedent component
  (\figrefsub{adm-match}{b}).\label{step:smy:sentmatch:align}
\item Annotate aligned source nodes indexed in
  step~\ref{step:smy:sentmatch:align} with alignment flows from \ds\
  component edges (\figrefsub{adm-match}{c}).
\item Associate the aligned node summary annotations for each respective
  sentence in the source component (\figrefsub{adm-match}{d}).
\item Create a sentence match candidate between the source and summary
  sentences (\figrefsub{adm-match}{e}).
\item Sort the source sentences by the sum of the flow from each summary
  sentence.\label{step:smy:sentmatch:srt}
\item Match sentences based on the flow from each summary to source
  sentence.\label{step:smy:sentmatch:match}
\item All remaining unmatched note antecedent sentences are given a stub
  \texttt{no-section} label.
\end{zlpackedenum}

Once the source sentences are paired with distributions of summary sentences by
flow in step~\ref{step:smy:sentmatch:srt} each source sentence is matched with
zero or more summary sentences.  In step~\ref{step:smy:sentmatch:match}, a
source sentence is matched with the summary sentence that has the maximum flow
determined by the minimum sentence flow hyperparameter.  The matched summary
sentence is then eliminated as a candidate for matching with any other source
sentence and the source sentences are tagged with the section of the matched
summary sentence.  Upon completion, antecedent sentences are tagged with the
\ds\ section to which it should be added.  For example, a sentence in a
radiology antecedent note marked with \texttt{Brief Hospital Course}
would be added to the \ds\ under that section during the generation
process.

\subsect[smy:srcsecds]{Learn Section Types}

We refer to the set of notes that were successfully aligned as the Source
Section Dataset.  The Sentence Matching Algorithm just described in
\secref{smy:sentmatch} was used to automatically pair sentences from note
antecedents to \dss\ of this dataset.  Each sentence pair of each admission
graph was used to train a section-assignment model (Source Section Model) using
alignment-induced labels.  The note counts by categories are given in
note-antecedent sentences from alignment-derived supervision, enabling
section-aware assembly with traceable evidence links.

\matchedSentSectionsTab[t] % -- Matched Sentence Sections

This dataset was automatically generated from \calamr\ outputs and was used as
weak supervision rather than as manually curated ground truth. We did not
perform a separate manual audit of these intermediate alignments/labels in this
study; instead, we validated the pipeline at the level of its intended clinical
output via human evaluation of the generated, multi-section discharge
summaries, assessing section-wise correctness and overall readability
\tabsee{informalEvalTab}.

The selected \ds\ sections, excluding the \texttt{no-section} label (a stub
used for unclassified sentences), were based on those that were considered most
necessary and beneficial for summarization by a physician authoring the note,
by a clinical informatics fellow and a 4\textsuperscript{th} year medical
student.  The physician-selected \ds\ sections and their counts are given in
\tabref{matchedSentSectionsTab}.  Most notable is the imbalance between the
section labels and \texttt{no-section} label.  This high disparity leads to a
terse generated \ds, which is explained further in \secref{smy:res}.
However, the \texttt{no-section} label preserves faithfulness by preventing
forced assignment when no aligned evidence exists.

Once the Sentence Matching Algorithm was used to assign labels to source
sentences \seesec{smy:sentmatch} a \ac{bilstm} was trained to learn the \ds\
section type of each note antecedent source sentence.  A section, such as
\textit{Hospital Course}, was a label predicted by the model indicating that not
only should the sentence be added, but to which section in the \ds\ to add it.
A label of \texttt{no-section} means the sentence is to be discarded.

A \ac{bilstm}~\cite{\bilstmct} was used for learning the sentence section
classification.  The \gatortron~\cite{\gatortronct} clinical embeddings, the
note antecedent's note category, and the section type were used as input
features to the model.  Because of the data input size~\seesec{smy:sentmatch} the
model's static embeddings were used in place of fine-tuning.  A fully connected
linear layer was added between the \ac{bilstm} and the output layer.  The
\ac{bilstm} layer had a hidden size of 500 parameters, a dropout of $p = 0.15$,
a learning rate of $5 \times 10^{-4}$ and used gradient clipping.  The model was
set to train for 30 epochs and converged at 24 epochs.

\subsect[smy:method:gen]{Generate Discharge Summary}

The \dss\ were generated using the Source Section
Model~\seesec{smy:srcsecds} trained on the Source Section Dataset.  The
note antecedents\ of the Source Section Dataset's test set were used as input
to the Source Section Model.  Sentences were added to the predicted section in
the generated \ds\ or discarded if the \texttt{no-section} label was predicted.

The \uicds\ was used as a development set by tuning the \acs{calamr}
$k$\textsuperscript{th} order neighbor set hyperparameter
(\ensuremath{\mathbf{\Lambda}}) to include more network neighborhood semantic
information.  The minimum sentence flow hyperparameter
(\ensuremath{\mathbf{\mu_{s}}}) was also adjusted to increase the output to 248
aligned admissions with higher quality.

The \mimic\ trained summarization model yielded 133 automatically generated
\dss\ and the \uicds\ model generated five.  The alignment challenges described
in \secref{smy:sentmatch}, such as missing \dss\ and GPU memory constraints,
show the difficulty of hospitalization summarization.  Further discussion of
these challenges are described in \secref{smy:res}.

% \dumpnote{ds-gold-sym-uic}{Gold Discharge Summary}{%
%   The physician hand written gold \uihealth\ de-identified discharge summary.
%   There were no modifications to this text other than adding bold type for the
%   headers and underlines for redacted text for readability.}

\begin{figure*}[h!]
  \centering
\begin{lstlisting}[style=zzdsstyle]
Discharge Summary by (*@\lstredact{[**Doctor First Name**]}@*) (*@\lstredact{[**Doctor Last Name**]}@*), MD at (*@\lstredact{[**Date**]}@*) 6:00 AM

Author: (*@\lstredact{[**Doctor First Name**]}@*) (*@\lstredact{[**Doctor Last Name**]}@*), MD Service: Neuro Critical Care Author Type: Resident
Filed: (*@\lstredact{[**Date**]}@*)  6:09 PM Date of Service: (*@\lstredact{[**Date**]}@*)  6:00 AM Status: Attested
Editor: (*@\lstredact{[**Doctor First Name**]}@*) (*@\lstredact{[**Doctor Last Name**]}@*), MD (Resident) Cosigner: (*@\lstredact{[**Doctor First Name**]}@*) (*@\lstredact{[**Doctor Last Name**]}@*), MD at (*@\lstredact{[**Date**]}@*) 12:45 PM
Attestation signed by (*@\lstredact{[**Doctor First Name**]}@*) (*@\lstredact{[**Doctor Last Name**]}@*), MD at (*@\lstredact{[**Date**]}@*) 12:45 PM  Stroke attending: I have reviewed the above discharge summary and agree with the assessment.

Pt eloped before I could staff the pt.

(*@\lstredact{[**Hospital Name**]}@*)
Discharge Summary

Patient: (*@\lstredact{[**First Name**]}@*) (*@\lstredact{[**Last Name**]}@*)
Admission Date: (*@\lstredact{[**Date**]}@*)
Discharge Date: (*@\lstredact{[**Date**]}@*)
Discharge Disposition: Left Against Medical Advice
Discharge Service: Stroke
Discharge Attending: (*@\lstredact{[**Doctor First Name**]}@*) (*@\lstredact{[**Doctor Last Name**]}@*), MD
Primary Diagnosis: Acute R medial temporal and internal capsule/thalamic stroke

(*@\lstheader{Other Active Diagnoses}@*)
Diagnosis Date Noted POA
- Troponin level elevated (*@\lstredact{[**Date**]}@*) Yes
Priority: High
- Stroke (CMS/HCC) (*@\lstredact{[**Date**]}@*) Yes

(*@\lstheader{Hospital Course}@*)
(*@\lstheader{HPI}@*):(*@\lstredact{[**First Name**]}@*) (*@\lstredact{[**Last Name**]}@*) 47 y.o.PMH HTN, HLD, previous stroke (*@\lstredact{[**Date**]}@*) (R-MCA watershed), presenting to ED with L sided weakness, LLE numbness, and blurry vision b/l. Patient reports symptoms started acutely on Friday (*@\lstredact{[**Date**]}@*) around 4pm while driving causing him to have to pull over. He decided to try to sleep it off. After waking up the following morning with no improvement, he went to (*@\lstredact{[**Hospital Name**]}@*) ED where he was seen by neurology, but left AMA as he felt he was being asked the same questions repeatedly and nothing was getting done. CTH at (*@\lstredact{[**Hospital Name**]}@*) was without ICH, reportedly showed wedge shaped hypodensity in frontal lobe likely from chronic infarct. The patient reports that the left sided weakness has improved somewhat today, but he still endorses b/l blurry vision with constant white floaters, as well as numbness/tingling in his LLE. NIHSS is 2 (LUQ quadrantopia, LLE numbness). CTA Hmultifocal narrowing of b/l ACAs as well as R M1 focal narrowing.B w/oacute ischemia in medial R temporal lobe involving the posterior limb of R internal capsule as well as possibly the thalamus, and redemonstrating old R frontal ACA-MCA watershed infarct and old L occipital cortical infarct. Patient to be admitted to stepdown under stroke service.

Patient with poorly controlled HTN and HLD, not takingthe past 5 monthshe says he had trouble getting primary care appointmentprescription renewals. SBP gets up to 200s per patient. Labs today significant for Troponin of 0.77-0.62. EKG showing T inversions in V5 and V6. Cardiology consulted. Patient denying CP at this time. Cardio recommending trending EKG/trop until downtrend and ordering Echo.

Patient appeared to have left before he was evaluated (*@\lstredact{[**Date**]}@*) AM.

Pertinent Physical Exam At Time of Discharge
Physical Exam
PATIENT NOT EXAMINED PRIOR TO DISCHARGE

Test Results Pending At Discharge

(*@\lstheader{Discharge Medications}@*)

No medications have been prescribed.

Issues Requiring Follow-Up
- Patient not evaluated prior to discharge
Outpatient Follow-Up Appointments
No future appointments.
Referrals
No orders of the defined types were placed in this encounter.

(*@\lstheader{Completed Consults:}@*)
Consults Ordered This Encounter
Procedures
- Inpatient consult to General Neurology
- Inpatient consult to Cardiology
\end{lstlisting}%
  \caption[Gold Discharge Summary]{\textsc{Gold Discharge Summary.}
  The physician hand written gold \uihealth\ de-identified discharge summary.
  There were no modifications to this text other than adding bold type for the
  headers and underlines for redacted text for readability.}
  \label{fig:ds-gold-sym-uic}%
\end{figure*}

\sect[smy:method:eval]{Evaluation Setup}

F1, precision and recall metrics were used to evaluate the source section
model.   For summarization, we did not directly compare
with published baselines due to differing evaluation pipelines.  Instead, our
focus was evaluation for provenance for the joint purposes of faithfulness and
traceability.

Less than half of the \ehr\ note text was represented in the
\ds~\cite{adamsWhatSummaryLaying2021,\pldsprovct}.  For additional context, we
reference published \ds\ baselines of \citet{zhuLeveragingSummaryGuidance2023}.
However, we do not treat their reported scores as directly comparable due to
differences in input scope, target definition, and evaluation pipelines.
Encounter--level summarization is intrinsically difficult because \ds\ content
often  lexically differs from note antecedent text despite being semantically
related.  This motivated cross-document provenance-centered evaluation that
prioritized evidence traceability over surface overlap.

\subsect{Limitation of Automatic Evaluation Metrics}

Automatic metrics on generated summaries, such as \rougename~\cite{\rougect},
\acs{bleu}~\cite{\bleuct} and \acs{bertscore}~\cite{\bertscorect}, are of
little help with such a large disjoint set of textual documents.  Still,
we compared the \ehr\ records with the discharge summary,
by concatenating the note antecedents for comparison.
Automatic overlap metrics were uniformly low in this encounter-level setting.
Specifically, \rougename{1} was \avmRone, \rougename{2} was \avmRtwo, and
\rougename{L} was \avmRL; \bleu\ was \avmBleu. These values reflected the
substantial lexical mismatch between antecedent notes and discharge summaries.
Accordingly, we treat \rougename\ and \bleu\ as a coarse
reference and emphasize human review based on traceable evidence links to
source spans.

Because encounter-level discharge summaries often differed
lexically from antecedent notes, and because published baselines and ablations
varied substantially in input scope and target definition (and typically did
not enforce provenance), we report provenance-centered results and refer the
reader to baseline papers by
\citet{lewisBARTDenoisingSequencetoSequence2020} and
\citet{zhangPEGASUSPretrainingExtracted2020} for non-comparable context.  For
these reasons we believe human evaluation is appropriate for judging the
effectiveness of generated documentation given the depth, complexity and
technical jargon found in clinical notes.

This evaluation of \dss\ on the Source Section Dataset's test set was evaluated
by a clinical informatics fellow and a 4\textsuperscript{th} year medical
student.  Each generated \ds\ was ranked using a Likert
scale~\cite{likertTechniqueMeasurementAttitudes1932} as an integer value
ranking in the range 1 to 5 with five as the highest on the following
questions:
\begin{zlenumerateinline}
\item preference ``do you prefer the generated summary''
\item readability ``of the data in the generated summary, how readable is it''
\item correctness ``of the data that is in the generated summary, how correct is it''
\item complete ``how complete is the generated summary''
\item sections ``of the data in the summary, how well is it sectioned''
\end{zlenumerateinline}

\subsect[smy:res:parsereval]{Parser Assessment}

The choice of \amr\ parser has the potential of greatly affecting performance.
Poor performance by the parser leads to error propagation because of semantic
use of its output and because it is one of the first components of the
pipeline.  Given the importance of the \amr\ parser, we evaluated several
parsers for correctness for the clinical domain.

Two \amr\ parsers were fine-tuned on the \ac{bioamrcorp} and then judged for
correctness by the clinical reviewers.  Nine graphs were scored with the T5
amrlib trained parser~\cite{jascobAmrlib2022} and another nine scored with the
Gsii parser~\cite{\gsiiparct}.  The qualitative analysis of the graphs by the
clinical reviewers led to the conclusion that the generated graphs were
insufficient.  An evaluation of \amr\ graphs created by the \ac{thyme}
demonstrated significantly improved results.  This analysis gave sufficient
motivation to use the \ac{thyme} for the remainder of the experiments.

\dsGenSummaryTab[t]{\tabsize}

\informalEvalTab[b]{\tabsize}

\sect[smy:res]{Results}

% \dumpnote{ds-gen-sym-uic}{Generated Discharge Summary}{ A \acl{uihealth}
%   de-identified automatically verbatim generated discharge summary as plain
%   text. There were no modifications to this text other than adding bold type
%   for the headers and underlines for redacted text for readability.}

\begin{figure*}[h!]
  \centering
\begin{lstlisting}[style=zzdsstyle]
(*@\lstheader{History of present illness}@*):
9:36 AM Status: Attested Editor: (*@\lstredact{[**Doctor First Name**]}@*) (*@\lstredact{[**Doctor Last Name**]}@*), MD (Resident) Related Notes: Original Note by (*@\lstredact{[**Doctor First Name**]}@*) (*@\lstredact{[**Doctor Last Name**]}@*), MD (Resident) filed at (*@\lstredact{[**Date**]}@*) 4:01 PM Cosigner: (*@\lstredact{[**Doctor First Name**]}@*) (*@\lstredact{[**Doctor Last Name**]}@*), MD at (*@\lstredact{[**Date**]}@*) 12:46 PM Consult Orders 1. Inpatient consult to General Neurology [(*@\lstredact{[**Correspondence ID**]}@*)] ordered by (*@\lstredact{[**Doctor Last Name**]}@*) (*@\lstredact{[**Doctor Last Name**]}@*), MD at (*@\lstredact{[**Date**]}@*) 0754 Attestation signed by (*@\lstredact{[**Doctor First Name**]}@*) (*@\lstredact{[**Doctor Last Name**]}@*), MD at (*@\lstredact{[**Date**]}@*) 12:46 PM Stroke Attending: Pt eloped prior to being seen. Blurred Vision and Extremity Weakness (*@\lstredact{[**First Name**]}@*) (*@\lstredact{[**Last Name**]}@*) is a 47 y.o. male with PMH HTN, HLD, previous stroke (*@\lstredact{[**Year**]}@*) (R ACA-MCA watershed), presenting to ED with L sided weakness, LLE numbness, and blurry vision b/ l. Patient reports symptoms started acutely on Friday (*@\lstredact{[**Date**]}@*) around 4 pm while driving causing him to have to pull over. He decided to try to sleep it off. After waking up the following morning with no improvement, he went to (*@\lstredact{[**Hospital Name**]}@*) ED where he was seen by neurology, but left AMA as he felt he was being asked the same questions repeatedly and nothing was getting done. CTH at (*@\lstredact{[**Hospital Name**]}@*) was without ICH, reportedly showed wedge shaped hypodensity in frontal lobe likely from chronic infarct. The patient reports that the left sided weakness has improved somewhat today, but he still endorses b/ l blurry vision with constant white floaters, as well as numbness/ tingling in his LLE. CTA Hwith multifocal narrowing of b/ l ACAs as well as R M1 focal narrowing. MRI B w/o showing acute ischemia in medial R temporal lobe involving the posterior limb of R internal capsule as well as possibly the thalamus, and redemonstrating old R frontal ACA-MCA watershed infarct and old L occipital cortical infarct. Patient to be admitted to stepdown under stroke service. Patient with poorly controlled HTN and HLD, not taking any medications for the past 5 months as he says he had trouble getting primary care appointment for prescription renewals.

(*@\lstheader{Physical examination}@*):
NIHSS is 2 (LUQ quadrantopia, LLE numbness). SBP gets up to 200s per patient. Labs today significant for Troponin of 0.77. EKG showing T inversions in V5 and V6. Cardiology consulted. Patient denying CP at this time. Prior stroke/ TIAs (date, description): R ACA-MCA waterhsed stroke in (*@\lstredact{[**Year**]}@*) per ED, worked up at (*@\lstredact{[**Hospital Name**]}@*), on DAPT Vascular risk factors: HTN, HLD, prior stroke Past Medical/ Surgical history: Past Medical History: Diagnosis Date Hypertension
\end{lstlisting}%
  \caption[Generated Discharge Summary]{\textsc{Generated Discharge Summary.}  %
    A \acl{uihealth} de-identified automatically verbatim generated discharge
    summary as plain text. There were no modifications to this text other than
    adding bold type for the headers and underlines for redacted text for
    readability.}\label{fig:ds-gen-sym-uic}%
\end{figure*}

The Source Section Model results are summarized in \tabref{dsGenSummaryTab}.
The weighted F1 score of 88.72 on the \mimic\ corpus shows good performance for
\ds\ section classification. However, the macro F1 was 20.41. The high weighted
F1 largely reflected the dominant \texttt{no-section} label, while macro F1
reflected difficulty on minority sections due to their sparsity.

These results indicate strong performance on frequent sections, supporting
section-aware assembly of evidence-linked (provenance) multi-section outputs
from note antecedents. Macro F1 should therefore be interpreted primarily as a
measure of minority-section difficulty in the presence of a dominant
\texttt{no-section} label.  The model trained on the \uicds\ shows lower
results.  This might be due to the higher rate of reentrancies as shown
in~\tabref{graphStatMergeTab} and discussed in~\secref{meth:align}.  The fact
that \mimic\ is a curated dataset is the most likely reason the results are
higher compared to \uicds, which is unmodified and contains \ac{hphi}.

\Tabref{informalEvalTab} shows the quantitative evaluation of 133 generated
\dss\ trained on the \mimic\ corpus.  The evaluation illuminates the difficulty
of the task and corroborates the low automated metrics between the gold \ehr\
note antecedents with the gold \ds.  The generated summaries achieve a perfect
correctness score (5), and an average readability; despite low scores on
sectioning, completeness and preference, this signals the promise of graph
methods with \amr\ for \textit{faithful} summarization.

The label imbalance in the Source Section Dataset might be attributed to the
sparsity of \calamr's alignments.  If this were the case, we could adjust the
hyperparameters\ of \calamr\ to produce more sentence matches.  However, the
lack of alignment could be justified by the lack of notes (other than those
from the \ac{icu} department) present in the \mimic\ corpus.  The misalignment
could also be attributed in cases where the physician writes from personal
experience with the patient that is otherwise lacking from the \ehr\ notes.

The \dss\ produced by the model trained on the \uicds~\tabsee{informalEvalTab}
show better completeness but slightly lower readability.  A higher sectioning
score was given to the \uicds\ despite the fact that the \medsecid\ model was
trained on \mimic.  This implies the \medsecid\ is able to section the \uicds\
notes or the Source Section Model is able to predict sections based on other
factors such as better alignments.  A gold \ds\ is given in
\figref{ds-gold-sym-uic} and its generated counter-part in
\figref{ds-gen-sym-uic}.

\sect[smy:conc]{Conclusions and Future Work}

We introduced an alignment-derived supervision strategy and section-aware
assembly for multi-section encounter--level discharge summarization. The
resulting output produced summaries with explicit sentence-level provenance
traceability to source spans, supporting faithful and auditable use in clinical
documentation. The Source Section Model showed promising results in assigning
\ds\ section labels to note-antecedent sentences using training data induced
from \calamr\ alignments.  These findings were further supported by physician
qualitative evaluations, which reported high correctness and reasonable
readability.

Shortcomings of the data used to train the models led to challenges that
affected performance and brought to light certain limitations.  In the case of
the \mimic\ data, the issue of omitted \ac{icu} notes~\cite{\dsprovct} led to
worse summarizations.  The summaries evaluated are faithful in that only
content from the source text is added to the summary.  They are traceable in
how each sentence can be traced back via the \calamr\ alignments.  Our early
experiments show promise despite the difficulty of the task.  Evaluation on a
broader generalizability dataset and ablations across alignment-quality,
section granularity, and long-context tasks, remain future work.

The utilization of a \acp{gnn} constructed from \calamr\ alignments could allow
for quality abstractive summaries (also left as future work).  We recommend a
modified version of a \ac{grnn}~\cite{\grnnct} that uses \calamr\ alignments to
train a graph induction model for summarization.  Leveraging the source
component could mitigate the need for a large training dataset since the model
would have more context for summarization.  The proposed modifications to the
\ac{grnn} would treat the source component as a Bayesian prior over the
graph-generation process using a \acs{lstm} cell for each iteration of the node
and edge insertion, much like the method of \citet{\gsiiparct}.

While \llms\ have great potential, hallucinations remain a limiting factor for
automatic generation of \dss.  We believe \amr\ graph alignment techniques
represent a promising frontier with the potential to generate accurate and
interpretable summaries for clinical applications.

\sect[sec:lim]{Limitations}%

Many of the admissions were too large for the constraints of the available
hardware when running the experiments.  Processing very large admissions (the
largest admission has 1,233 notes in the \mimic\ corpus) resulted in
performance issues given the immense scale of source text.  We leave
performance enhancements of the pipeline that would ameliorate these issues as
future work.

\sect{Acknowledgments}

This work was partially supported by award R01 CA225446 from the National
Institutes of Health (NIH) and by a postdoctoral award by the Center for Health
Equity using Machine Learning and Artificial Intelligence (CHEMA) at the
University of Illinois Chicago.  We thank Jon Cai and Professor Martha Palmer
at the University of Colorado Boulder for generously making the \ac{thyme}
available.

\clearpage
\zavenenddoc%

\end{document}